\documentclass{article}
\usepackage{spconf,amsmath,graphicx}
\usepackage{booktabs}
\usepackage{subcaption}
\usepackage{xcolor}
\usepackage{cite}
\usepackage{textcomp}
\usepackage{enumitem}
 
\title{An Ontology-aware framework for Audio event classification}
\name{Yiwei Sun$^{1,2}$ and Shabnam Ghaffarzadegan $^2$ \footnote{yus162@psu.edu, shabnam.ghaffarzadegan@us.bosch.com}}
\address{$^1$The Pennsylvania State University, State College, PA, USA \\ 
$^2$Bosch Research and Technology Center, Sunnyvale, CA, USA
}

\begin{document}
%
\maketitle

\begin{abstract}
Recent advancements in audio event classification often ignore the structure and relation between the label classes available as prior information. This structure can be defined by ontology and augmented in the classifier as a form of domain knowledge.
To capture such dependencies between the labels, we propose an ontology-aware neural network containing two components: feed-forward ontology layers and graph convolutional networks (GCN). The feed-forward ontology layers capture the intra-dependencies of labels between different levels of ontology. On the other hand, GCN mainly models inter-dependency structure of labels within an ontology level.
The framework is evaluated on two benchmark datasets for single-label and multi-label audio event classification tasks.
The results demonstrate the proposed solutions efficacy to capture and explore the ontology relations and improve the classification performance. 
\end{abstract}
\begin{keywords}
Audio event classification, Ontology structure, Graph convolutional networks, Ontology layers. \let\thefootnote\relax\footnote{e-mail: yus162@psu.edu, shabnam.ghaffarzadegan@us.bosch.com}
\end{keywords}

\vspace{-0.3cm}
\section{Introduction} \label{intro}
\vspace{-0.1cm}
Developing machine listening system similar to human hearing ability to make sense of sounds is one of the growing areas of research~\cite{gemmeke2017audio,Takahashi2016,Zhuang,45611}. Humans are capable of understanding ambiguous and over-lapping sounds and categorizing them into abstract concepts. This enables human to disambiguate sounds and understand the environment. For example, humans can categorize \textit{shouting} and \textit{baby crying} as sub-categories of \textit{human sound}. Or they can categorize \textit{car horn} either as a general category of \textit{street sound} or a more specific category of \textit{car sound}. However, majority of the state-of-the-art works in machine listening, specifically in the filed of audio event classification, do not take advantage of these abstract concepts.



To bridge this gap between human and machine hearing abilities, we aim to augment audio event classification models with the ontology structure of the abstract sound categories. This structure is usually available as prior information in the form of common or general knowledge.
Ontology represents the formal structure of classes or types of objects within a domain. As a result, ontology enables the models to process data in the context of what is known. 

Incorporating the hierarchical relations in the form of ontology in audio domain is a non-trivial task since there is no intrinsic graph structure in the data. 
However if used successfully, it brings multiple benefits for audio event classification models.
It learns robust representations which can disambiguate audio classes that are acoustically similar but semantically different.
It can also classify audio events in more general descriptors in case of ambiguity in sub-classes. 
Finally, in case of multi-label classification task, it incorporates the possibilities of events co-occurring in real world.

There are very few works that incorporated ontology for audio event classification. For example, authors in~\cite{raj_ontology} presented two ontology-based neural networks: feed-forward ontological layer and Siamese neural network. Using both methods, authors showed improved performance for common sound event classification datasets.
Having said that, ontological information has been incorporated more in other domains such as computer vision~\cite{wang2018zero, chen2019multi, zhou2019improving} and natural language processing~\cite{inproceedings, wang2018cross}. 

In this work, we propose an end-to-end ontology-aware audio event classification model to capture the entities correlations in the ontology.
More specifically, we define a task-independent and a task-dependent ontology structures to incorporate in our models --
(1) \textit{semantic ontology}: defined based on semantic relationship between labels. For example, \textit{dog} belongs to the category \textit{animal} and \textit{dog barking} and \textit{whining} belong to the category \textit{dog}. This ontology is task independent and can be defined by human linguistic;
(2) \textit{context ontology}: defined based on the context of the task. For example in the context of street, \textit{music} and \textit{car engine} sounds might co-occur frequently. However, there is no semantic relationship or acoustic similarities between the two. This task dependent ontology can be extracted from annotated data and is usually not included in common ontology trees released with
datasets. 
Our proposed ontology-aware framework includes two components to model two aforementioned ontology structures: 1) feed-forward ontology layers to incorporate semantic ontology and 2) Graph Convolutional Networks (GCN)~\cite{kipf2016semi} to incorporate context ontology.

The main contributions of this paper are: 1) We study multi/single-label audio event classification problem and propose an end-to-end ontology-aware network. 2) We provide a novel and general framework to augment deep learning models with prior auxiliary information via feed-forward ontology layers and Graph Convolutional Networks. 3) We conduct extensive experiments on real-world datasets to demonstrate the effectiveness of the proposed method.

\vspace{-0.3cm}

\section{Methodology}
\vspace{-0.1cm}
In this section, we present the proposed framework on how to augment semantic and context ontologies into a deep learning model. 
As a result, the trained model is able to take into account label correlations rather than treating classes as a set of independent parameters to be learned. 
As mentioned before, we incorporate two different components in our solution to model ontology structures. Note that these two components are independent of the base model and can be added to any neural network frameworks. 

\begin{figure}[t!] 
\centering
\includegraphics[width=\columnwidth]{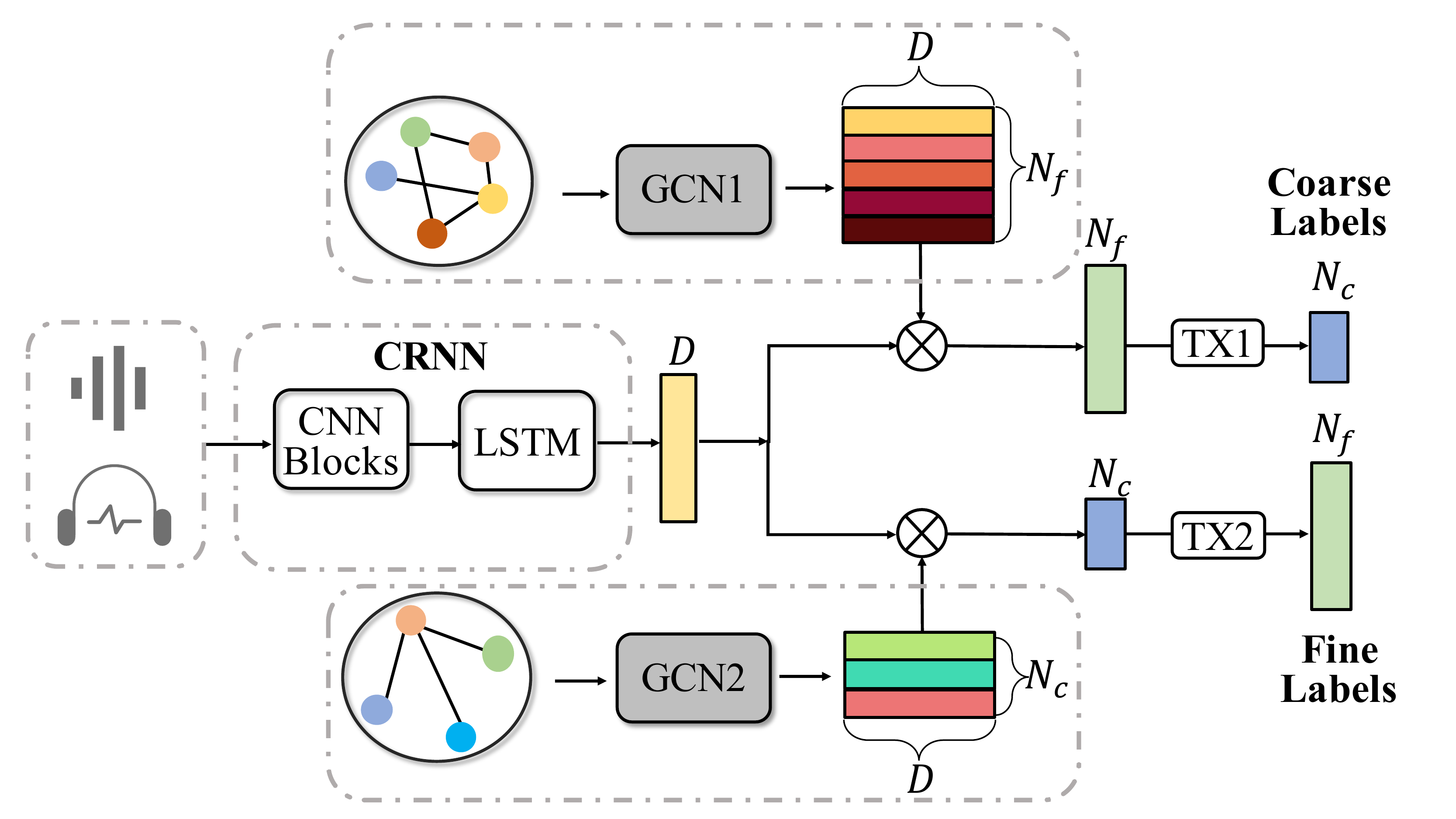}
\vskip -0.5em
\caption{The architecture of the ontology-aware model incorporating ontology feed-forward layers, $TX1$ and $TX2$, and Graph Convolutional Networks, $GCN1$ and $GCN2$.}
\vskip -0.8em
\label{diag1}
\end{figure}

\textbf{Feed-Forward Ontology Layers}: This layer is the first component in our solution that enables prediction of one level of hierarchy using the other levels embedding. Fig.~\ref{diag1} shows an example of a system with two levels of hierarchy, fine labels and coarse labels with two feed-forward ontology layers, $TX1$ and $TX2$.
As a result, ontology relations can be incorporated in all directions of the hierarchical levels. 
This is in contrast to the feed-forward ontology layer of~\cite{raj_ontology} which only allows one direction relation.

\textbf{Graph Convolutional Networks} \label{GCN}
The second component of our solution is Graph Convolutional Networks (GCNs).
Recently, GCNs have achieved immense success in capturing the underlying associations and correlations between entities and have been widely adapted to various domains such as computer vision~\cite{chen2019multi} and natural language processing~\cite{Yao2018GraphCN,Cai2017ACS}. However, to the best of our knowledge, GCNs have never been explored in audio domain. Using GCN in audio domain is a non-trivial task since there is no intrinsic graph structure in the data. Here, we propose audio GCNs to construct the label graph based on the auxiliary information and propagate associations and correlations in the ontology structure.

A GCN is a multi-layer graph neural network that generalizes the convolution operation from a grid data to graph data~\cite{bronstein2017geometric}. GCN learns nodes representation based on its own feature and neighbors feature. One layer GCN can capture information from its immediate neighbors. However, in multi-layer GCN information is propagated from larger neighbors.

Consider Graph $G$ with $n$ nodes. Let $X \in \mathcal{R}^{n\times m}$ be a matrix with features of $n$ nodes with $m$ dimensions. Let $A$ be the adjacency matrix of graph $G$ with its degree matrix $D$ defined as $D_{ii}=\sum_jA_{ij}$. The diagonal elements of $A$ are set to 1 due to self-loops. For multi-layer GCN, which is the result of stacking multiple single layer GCNs, $k$-dimensional node feature matrix $L^{(j+1)} \in \mathcal{R}^{n\times k}$ is calculated as:
\begin{equation}
    L^{(j+1)}=f(\hat{A}L^{(j)}W_j),
\end{equation}
where $j$ is the layer number,  $\hat{A}=D^{-\frac{1}{2}}AD^{-\frac{1}{2}}$ is the smoothed adjacency matrix, $W_j$ is the weight matrix, $f$ is the non-linear activation function such as ReLU, and $L^{(0)}=X$.
To build a GCN for audio data, we define number of nodes, initial embdeddings of each node and graph edges as: (1) We set number of nodes to the number of labels in the whole datasets; (2) The initial embeddings of each node is set to the word embedding of the labels. For the labels with more than 1 word, we use average of the embedding. For example, for the label \textit{small engine}, the final word embedding is the average of the word embeddings of \textit{small} and \textit{engine}; (3) Finally, we define graph edges based on labels co-occurrence.



\begin{figure}[t!] 
\centering
\includegraphics[width=0.95\columnwidth]{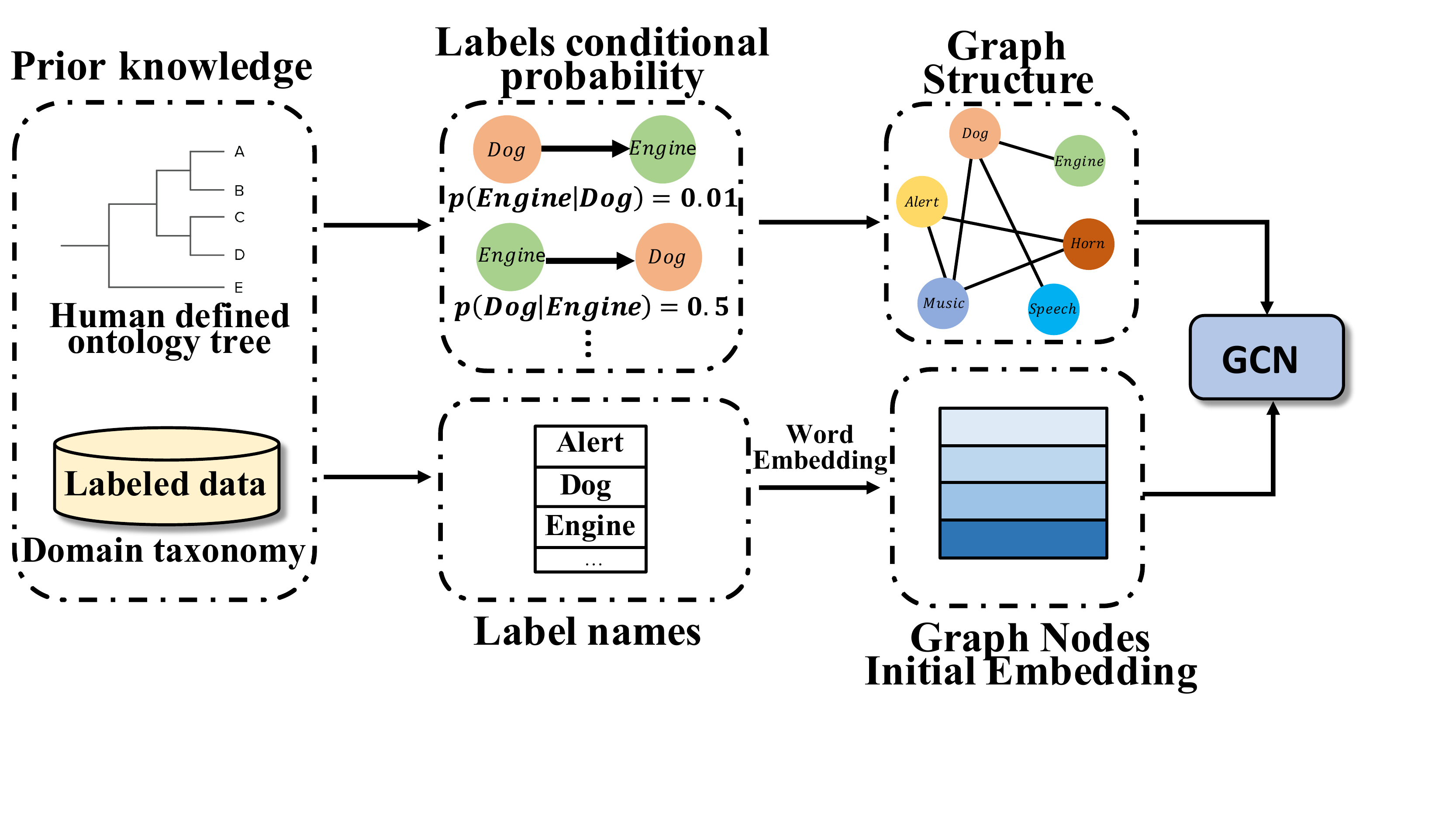} 
\vskip -2em
\caption{The framework of constructing audio Graph Convolutional Networks. Each node denotes a label and word embedding is used as initial node representation. Graph edges are built based on labels conditional probabilities.}
\vskip -1em
\label{diag2}
\end{figure}

Inspired by~\cite{chen2019multi}, this is calculated via conditional probability between the labels which can be extracted from data annotation.
Note that data annotation includes information from both semantic and context ontology.
A summary of different steps of building the audio GCN is presented in Fig.~\ref{diag2}.
More concretely, conditional probability between labels $L_i$ and $L_j$ is defined as  $P_{ij}=count_{i|j} / count_{j}$. Where $count_{i|j}$ is the count of appearance of $L_i$ when $L_j$ exists and $count_{j}$ is the total count of $L_j$ in the whole corpus.
Fig.~\ref{diag2} shows an example of calculating conditional probability of labels \textit{dog} and \textit{engine}.  
Based on~\cite{chen2019multi} to avoid over-fitting the adjacency matrix to the training data, we first binarize the adjacency matrix as:
\begin{equation}\label{eq1}
A_{ij}=
\begin{cases}
  0, & \text{if }P_{ij} < \tau\\    
  1, & \text{if }P_{ij}\geq \tau
\end{cases}
\end{equation}
Where $\tau$ is the hyper-parameter threshold level. Moreover, to avoid over-smoothing we re-weight the adjacency matrix as:
\begin{equation}\label{eq2}
A'_{ij}=
\begin{cases}
  p/\sum_{j=1}^{n}A_{ij}, & \text{if }i \neq j\\    
  1-p, & \text{if }i = j
\end{cases}
\end{equation}
Where $n$ is the number of nodes and $p$ is a hyper-parameter which defines the weight assigned to each node and its neighbors.
In our experiments we also built multiple graphs for each level of hierarchy in the ontology. In this case, number of nodes for each graph is equal to the number of labels at each level and the value of edges are calculated as explained before.


\textbf{Overall Network Architecture}: As shown in Fig.~\ref{diag1}, the final model includes a base network, feed-forward ontology layers and GCNs classifying coarse and fine labels. 
Note that the overall architecture is generalizable to more levels of hierarchy.
Here, convolutional neural network (CNNs) and Long-Short-Term-Memory (LSTM) are used as the base network.
We have two feed-forward ontology layers and two GCNs. GCNs model inter-dependency structure of each level and feed-forward ontology layers model intra-dependency between two levels of the hierarchy.
Concretely, $GCN1$ is built on the second level of the hierarchy with nodes corresponding to the fine labels. $GCN2$ is built on the first level of the hierarchy with each node corresponding to a coarse label.
We use this architecture for multi-label classification. In the case of single-label classification there is no context ontology. Hence, we build one GCN on the whole taxonomy with number of nodes equal to the fine and coarse labels combined replacing $GCN1$ and $GCN2$ in Fig.~\ref{diag1}. 
In the experiment section, we will compare the performance of single and multiple GCNs for multi-label classification task and use single GCN for the case of single-label classification.

\textbf{Network Training}: Two-step process is used to update the network parameters in Fig.~\ref{diag1}. In the \textit{first step}, we calculate the output of the LSTM network as common embedding for predicting fine and coarse labels. Next, the common embedding is multiplied by $GCN1$ output to predict the fine labels. Note that $GCN1$ output is its nodes feature matrix with the size $D\times N_f$. Where $D$ is a hyper-parameter of the graph and $N_f$ is the number of fine labels. We use these fine labels estimation to calculate fine loss. 
Next, the fine labels prediction is fed to $TX1$ to output coarse label prediction. We then calculate coarse loss using this estimations. The final loss of the first step is weighted sum of the two losses. Finally, we perform back-propagation to update the network parameters. 
In the \textit{second step}, we calculate the common embedding with the updates weights and multiply it with $GCN2$ output with the size of $D\times N_c$ in which $N_c$ is number of coarse labels. This operation results in coarse labels prediction which we use to calculate coarse loss. 
Next, the coarse labels prediction are passed through $TX2$ to output fine labels estimate. Similarly, we calculate the final loss of the second step as the weighted sum of the fine and coarse labels loss. 
We perform back-propagation again to update the
network parameters.
We continue this process till convergence. Experimentally, this two-step update process have shown superior results than one step update using the final coarse and fine label predictions. 

\vspace{-0.3cm}
\section{Experiments}
\vspace{-0.1cm}
In this section, we introduce the datasets and the experiment settings. Moreover, we conduct extensive experiments for audio event classification task and present results to demonstrate the effectiveness of the proposed methods.
\vspace{-0.5cm}

\subsection{Datasets}
\vspace{-0.2cm}
\begin{itemize}[leftmargin=*]
\item \textbf{DCASE 2019-task5 (D19T5)}~\cite{Bello2019sonyc}: 
This dataset is used for multi-label Urban Sound Tagging in DCASE 2019 challenge, Task 5. It is recorded by an acoustic sensor network in New York city and is annotated for 23 urban sound noise labels, fine-labels. There are 8 main categories as coarse labels. All recordings are 10 seconds single-channel 44.1 kHz, 16-bit wave format. 
\vspace{-0.3cm}
\item \textbf{Urban Sounds (US8K)}~\cite{Salamon:2014}:
This dataset contains real field recordings of 8732 labeled urban sound excerpts with duration of less than 4 seconds from 10 classes with only one label for each recording. All the recordings are single-channel 44.1 kHz, 16-bit wave format. The files are pre-sorted into ten folds. Similar to \cite{raj_ontology}, we use two levels ontology with 4 classes in coarse level and 10 classes in fine level. 
\vspace{-0.4cm}
\end{itemize}
\vspace{-0.4cm}
\subsection{Experimental Setup}
\vspace{-0.2cm}
\textbf{Audio features}: 
Similar to \cite{kong2019cross}, all audio files are re-sampled to 32 kHz and audio samples are represented by log-Mel spectrogram with 64 Mel bins, a window size of 1024 samples, hop size of 500 samples, and  cut-off frequencies of 50 Hz to 14 kHz. 
For D19T5 data, the 10 seconds excerpts are used as the fixed input size. For US8K data, input size is set to 4 seconds and samples are zero-padded if necessary.

\noindent\textbf{Network Architecture}:
We use 8 layers CNNs to extract audio embeddings as used in \cite{kong2019cross}. We then add 1 layer LSTM network with the output size of $512$ to incorporate the sequential nature of audio data. The feed-forward ontology layers have one layer with \textit{sigmoid} non-linearity. Both GCNs contains 2 layers networks with the first layer of size (300$\times$400) and the second layer of size (400$\times$512). \textit{LeakyReLU}~\cite{Maas2013RectifierNI} with the negative slope of 0.2 is used as the non-linearity between the GCN layers and \textit{sigmoid} is used as the final non-linearity of the GCN. For the graph initial node representations, we extract 300 dimensional GloVe embeddings \cite{pennington2014glove} trained on the Wikipedia dataset. For the correlation matrix in Eq.~\ref{eq2}, we set $p$ to 0.2.
Stochastic gradient descent is used for network optimization, and binary cross entropy as the loss function. Learning rate is set to 0.001 for D19T5 and 0.0001 for US8K. The network is trained for 8000 iterations for D19T5 dataset and 20K iterations for US8K dataset. 
Threshold value $\tau=0.2$ is chosen to binarize the adjacency matrix in Eq. \ref{eq1}. Note that the hyper parameters are chosen experimentally based on the validation sets.
We implemented our network in Pytorch.
\begin{table}[t!]
\caption{Evaluation results on D19T5 dataset. FF: Feed-Forward, Mi: Micro, Ma: Macro.}
\label{dcase-tabel}
\vskip -1em
\centering
\resizebox{1\columnwidth}{!}{%
\centering
\begin{tabular}{ccccccc}
\toprule
          & \multicolumn{2}{c}{Fine-level } & \multicolumn{2}{c}{Coarse-level } \\
          \cmidrule(r){2-3}
            \cmidrule(r){4-5}
Methods & Mi AUPRC   & Ma AUPRC & Mi AUPRC & Ma AUPRC \\  \midrule
Baseline~\cite{Bello2019sonyc} &0.672  &  0.428 &  0.743  &  0.530\\
CNN9-avg~\cite{kong2019cross} & 0.672 & 0.433 & 0.782 &  0.628 \\
Our baseline &0.684 & 0.449&0.807&0.632\\
1 graph wo/ FF ontology  &0.678  &0.447 &0.802 &0.579  \\
1 graph w/ FF ontology&0.702  & 0.505 &0.813 &0.633\\
2 graphs wo/ FF ontology  &0.703  &\textbf{0.517} &0.819&\textbf{0.635} \\
2 graphs w/ FF ontology &\textbf{0.715}  &0.510 &\textbf{0.823} &0.625\\
\bottomrule
\end{tabular}}
\vskip -1em
\end{table}

\vspace{-0.2cm}
\subsection{Experimental results}
\vspace{-0.2cm}
In this section, we compare the proposed solutions to baseline methods on D19T5 for multi-label audio event classification task and US8K for single-label classification. 
We  compute  Micro and Macro Area Under Precision and Recall Curve (AUPRC) as metrics for D19T5 dataset as used in DCASE 2019 challenge Task 5. For US8K data, we use Macro and Micro F1 scores.

Table~\ref{dcase-tabel} shows evaluation results for D19T5 dataset for both coarse and fine level predictions. Baseline \cite{Bello2019sonyc} is the DCASE 2019 challenge Task 5 baseline system based on multi-label logistic regression model with VGGish embeddings \cite{45611}. The second row CNN9-avg \cite{kong2019cross} is another baseline on D19T5 data based on 9-layer CNN architecture with average pooling and Log-mel spectrogram input. As explained previously, our baseline model adopts the CNN9-avg architecture and adds an LSTM layer to incorporate sequential nature of the audio data. This addition shows improvement in both fine and coarse levels in Table~\ref{dcase-tabel}.  
Next, we compare the proposed ontology-based solution in 4 different settings to investigate the effect of each component of our solution. 
First, we use one GCN trained on the whole taxonomy with and without feed-forward ontology layers.
Based on the experiments, feed-forward ontology layers play an important role in achieving a superior performance compare to our baseline solution. 
The network with feed-forward ontology layers gets relative improvements of 2.6\% Micro AUPRC, 12.5\% Macro AUPRC for fine-level prediction and similar performance for coarse-level prediction. 

In the next experiment, we use two GCNs for fine and coarse levels with and without ontology layers. Separately modeling fine and coarse level taxonomies enables the GCNs to focus independently on the inter-dependency among labels in each levels. While allowing the feed-forward ontology layers to address intra-dependencies of the labels between the levels.
As shown in the table, models with two GCNs outperform the ones with only one GCN. 
Similar to the previous experiment the best performance is achieved when two GCNs and ontology layers are used together.
As shown in the Table~\ref{dcase-tabel}, this network achieves relative improvements to our baseline with 4.5\% Micro AUPRC, 13.6\% Macro AUPRC for fine-level prediction and 2.0\% Micro AUPRC and similar Macro AUPRC for coarse-level prediction.
We are ranked among top three solutions in the DCASE 2019 Task 5 challenge for both fine and coarse level predictions.

\begin{table}[t!]
\caption{Evaluation results on US8K dataset. FF: Feed-Forward, Mi: Micro, Ma: Macro.}\label{urbansound_results}
\vskip -1em
\centering   
\resizebox{1\columnwidth}{!}{%
\begin{tabular}{ccccccc}
\toprule
          & \multicolumn{2}{c}{Fine-level } & \multicolumn{2}{c}{Coarse-level } \\
          \cmidrule(r){2-3}
            \cmidrule(r){4-5}
Metrics & Ma F1 & Mi F1 & Ma F1 & Mi F1\\  \midrule
Our Baseline~\cite{raj_ontology} & 0.848& 0.832& 0.873& 0.873 \\
1 graph wo/ FF ontology  & 0.841 & 0.826&\textbf{0.890}&\textbf{0.890}\\
1 graph w/ FF ontology & \textbf{0.883} & \textbf{0.873} & 0.880&0.879\\
\bottomrule
\end{tabular}}

\vskip -1em
\end{table}

\begin{figure}[htbp]
\centering
\includegraphics[width=\columnwidth]{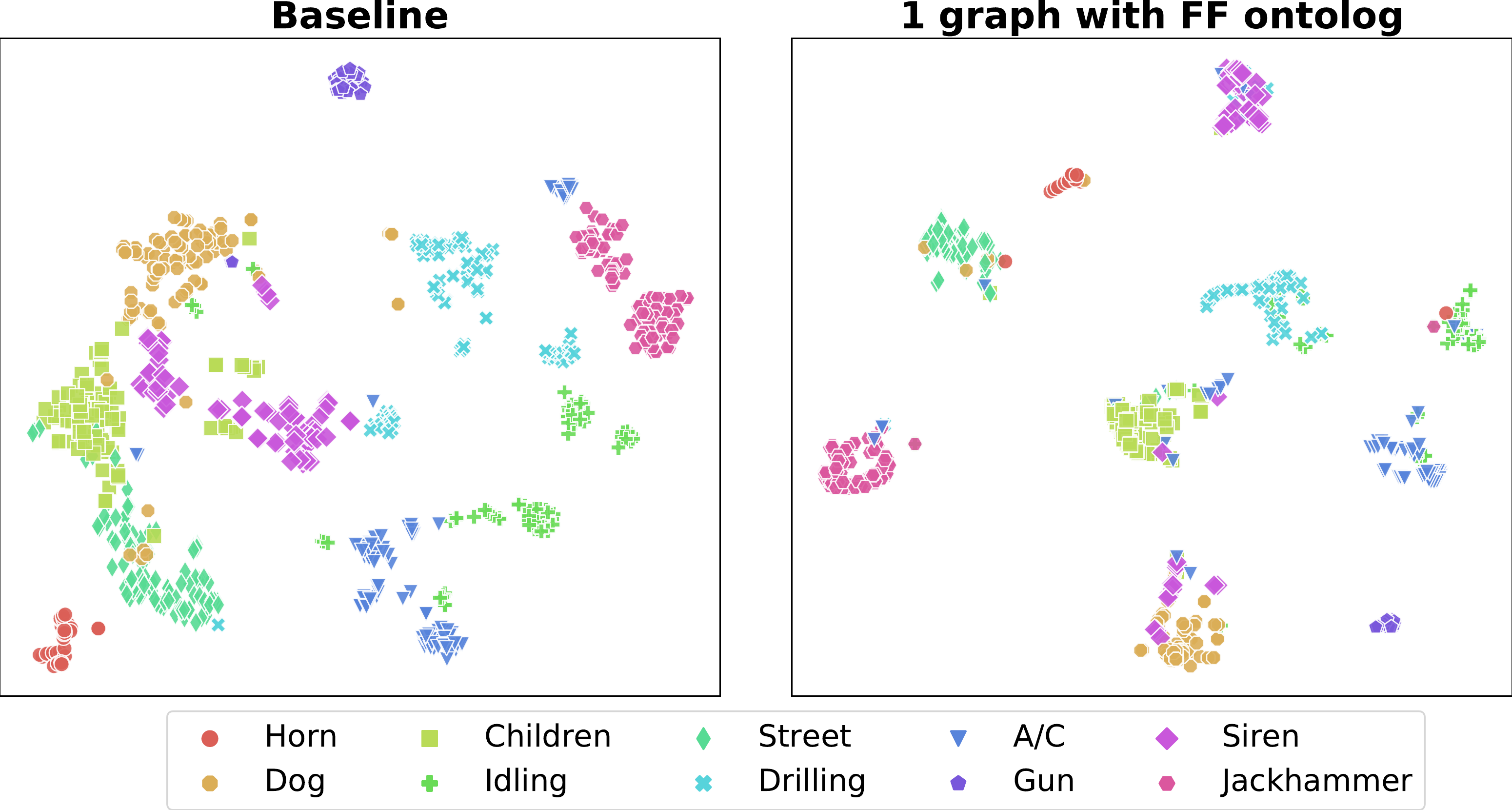} 
\vskip -0.8em
\caption{The t-SNE plots of the samples from US8K dataset in 10 fine classes for baseline and ontology-aware models.}
\vskip -1.2em
\label{tsne}
\end{figure}

In the next experiment, we use US8K dataset to investigate the effectiveness of the proposed solution for single-label classification. As mentioned in section~\ref{GCN}, there is no context ontology in a single-label classification. Hence, one GCN trained on the whole taxonomy is used as $GCN1$ and $GCN2$ in Fig.~\ref{diag1}. Table~\ref{urbansound_results} shows the classification results average on ten folds cross-validation using baseline, one graph with and without ontology layers. Similar to the results in D19T5 dataset, the architecture with feed-forward ontology layers outperform the one without. This model can achieve relative improvements of 4.1\% Macro F1 and 4.9\% Micro F1 for fine level and similar results for coarse level prediction. 
Finally, Fig.~\ref{tsne} demonstrates t-SNE plot with perplexity of 40 for the fine level classes in US8K. We observe a better grouping using the proposed ontology-aware model compared to baseline. 
In closing, our results confirm the effectiveness of different components of the proposed solution specially for fine level classification with more complicated correlations between the labels. We believe the two proposed components can be used in variety of deep learning architectures and domains other than audio.

\vspace{-0.3cm}
\section{Conclusion}
\vspace{-0.1cm}
In this paper, we proposed an ontology-based audio event classification to augment label dependencies in the model. The solution contains two components of feed-forward ontology layers and Graph convolutional networks. Using these elements, we model two ontology structures naming semantic ontology and context ontology. The results validated the effectiveness and importance of each component for multi-label and single-label audio event classifications. For future works, we would like to analyse the generalizability of the proposed framework to other neural network architectures and domains.


\bibliographystyle{IEEEbib}
\bibliography{reference}

\end{document}